\documentclass[a4paper]{article}

\usepackage{INTERSPEECH2016}

\usepackage{graphicx}
\usepackage{amssymb,amsmath,bm}
\usepackage{textcomp}
\usepackage{enumitem}

\def\bit{\begin{itemize}[noitemsep,nolistsep]}
\def\eit{\end{itemize}}
\def\it{\item}

\def\ben{\begin{enumerate}}
\def\een{\end{enumerate}}

\ninept

\title{Advances in Very Deep Convolutional Neural Networks for LVCSR}

\makeatletter
\def\name#1{\gdef\@name{#1\\}}
\makeatother \name{{\em Tom Sercu, Vaibhava Goel}}

\address{ Multimodal Algorithms and Engines Group, IBM T.J Watson Research Center, USA\\
  {\small \tt tsercu@us.ibm.com, vgoel@us.ibm.com}
}

\begin{document}

  \maketitle

\begin{abstract}
Very deep CNNs with small $3\times3$ kernels have recently been shown to achieve
very strong performance as acoustic models in hybrid NN-HMM speech recognition systems.
In this paper we investigate how to efficiently scale these models to larger datasets.
Specifically, we address the design choice of pooling and padding along the time dimension 
which renders convolutional evaluation of sequences highly inefficient.
We propose a new CNN design without timepadding and without timepooling,
which is slightly suboptimal for accuracy, but has two significant advantages:
it enables sequence training and deployment
by allowing efficient convolutional evaluation of full utterances, 
and, it allows for batch normalization to be straightforwardly adopted to CNNs on sequence data.
Through batch normalization, we recover the lost peformance from removing the time-pooling,
while keeping the benefit of efficient convolutional evaluation.

We demonstrate the performance of our models both on larger scale data than before, 
and after sequence training.
Our very deep CNN model sequence trained on the 2000h switchboard dataset
obtains 9.4 word error rate on the Hub5 test-set, matching with a single model the performance
of the 2015 IBM system combination, which was the previous best published result.

\end{abstract}

\section{Introduction}
We present advances and results on using very deep convolutional networks (CNNs)
as acoustic model for Large Vocabulary Continuous Speech Recognition (LVCSR), extending our earlier work \cite{sercu2015very}.

In \cite{sercu2015very}, we introduced very deep convolutional network architectures to LVCSR,
in the hybrid NN-HMM framework: the input to our CNN is a window of $(1+2\text{ctx})$ frames,
each frame a 40-D logmel feature vector,
as output we produce a vector of probabilities over CD~states for the center frame of the window.
We presented strong results on the Babel low-resource ASR task \cite{iarpababel}
and the 300-hour switchboard-1 dataset (SWB-1) 
after cross-entropy training only.
The very deep convolutional networks are inspired by the ``VGG Net'' architecture introduced in \cite{simonyan2014very}
for the 2014 ImageNet classification challenge.
The central idea of VGG networks is to replace layers with large convolutional kernels 
by a stack of layers with small $3\times3$ kernels.
This way, the same receptive field is created with less parameters and more nonlinearity.
Furthermore, by applying zero-padding throughout and only reducing spatial resolution through strided pooling, 
the networks in \cite{simonyan2014very} were simple and elegant. We followed this design
in the acoustic model CNNs \cite{sercu2015very}.

However, an important design choice remained unexplored in \cite{sercu2015very}: 
the accuracy and computational efficiency impact of time-pooling and time-padding
(i.e. zero-padding at the borders along the time dimension).
The networks from \cite{sercu2015very} pool in time with stride 2 on the higher layers of the network,
and applied time-padding throughout. This allowed for the elegant design analogous to the VGG networks
for computer vision. 
However, it leads to inefficient convolution over full sequences, making
sequence training infesible on large data sets. 
In this paper, we explore alternatives to that design choice.
Most importantly, we focus on designs that allow for efficient convolutional evaluation of full utterances,
which is essential during sequence training and test time (or deployment in a production system).

The key contributions of this paper are:
\bit
\it We demonstrate that convolutional models without time-pooling or time-padding, while being
    slightly suboptimal in recognition accuracy, allow for fast convolutional evaluation over sequences
    and enable sequence training over large data sets (sections \ref{sec:pooling} and \ref{sec:efficient}).
\it We demonstrate the merit of batch normalization (BN) \cite{ioffe2015batch} 
    for CNN acoustic models (section \ref{sec:bn}).
    BN is a technique to accelerate training and improve generalization by normalizing the internal representations inside the network.
    We show that in order to use batch normalization for CNNs during sequence training,
    it is important to train on several utterances at the same time.
    Since this is feasible for the efficient architectures only, batch normalization gives
    us essentially a way to compensate the lost performance from following the no time-padding design principle.
\it We present accuracy results of the very deep networks' performance after CE and sequence training on the 
    full SWB (2000h) corpus, achieving a 9.4 WER on Hub5 (1.0 WER better than the classical CNN).
\eit
  
\section{Exploring pooling and padding in time}
\label{sec:pooling}

\begin{figure*}[ht]
    \vspace{-1em}
    \centering
    \includegraphics[width=0.88\linewidth]{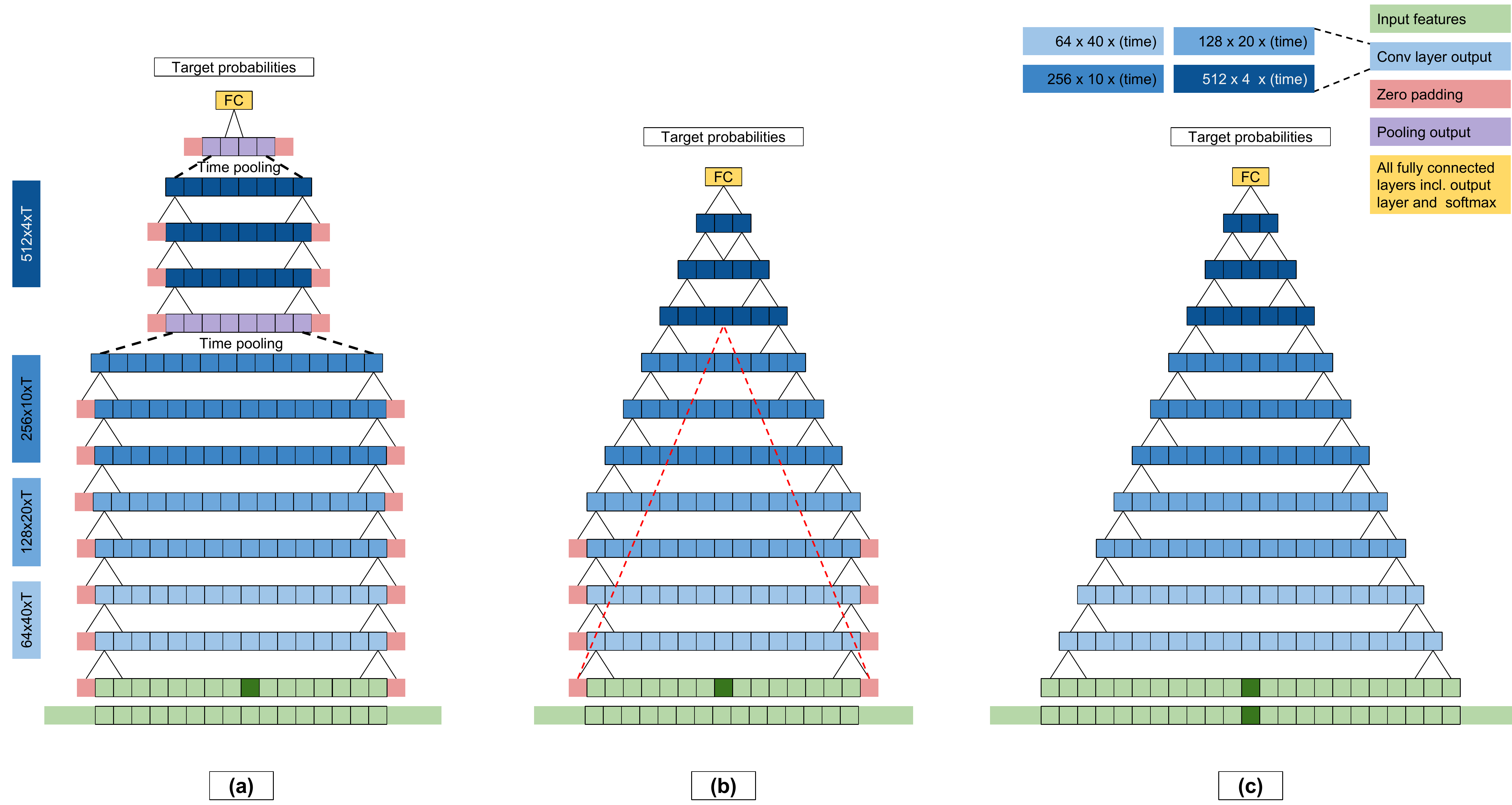}
    \vspace{-1em}
    \caption{Different versions of the 10-layer CNN from \cite{sercu2015very}. 
       The time-dimension is shown horizontally, CNN depth is shown vertically, the frequency dimension
       and number of feature maps are indicated with color shades.
       Pooling in frequency is implicitly understood at transitions between color shades.
       (a) This is the original (WDX) architecture from \cite{sercu2015very}, 
         starting from a 16-frame window.
       (b) Here we do not pool in time, but rather we leave out time-padding on the top layers.
            This reduces the size along the time-direction from 15 (ctx~$=7$) to 
            size 3, using the 6 highest convolutional layers.
       (c) If we want to remove the time-padding and pooling altogether, we need to
           start with a larger context window (ctx$=11$).
           This architecture is the only one that allows for efficient convolutional evaluation of full utterances
           (section \ref{sec:efficient}) and batch normalization (section \ref{sec:bn}).
       }
    \vspace{-0.3em}
    \label{fig:WDXpWDXnpWDX}
\end{figure*}

Pooling with stride is an essential element of CNNs in computer vision,
as the downsampling reduces the grid size while
building invariance to local geometric transformations.
In acoustic CNNs, pooling can be readily applied along the frequency dimension,
which can help build invariance against small spectral variation.
In our deepest 10-layer CNN in \cite{sercu2015very}, 
we reduce the frequency-dimension size from 40 to 20, 10, 4, 2 through pooling
after the second, fourth, seventh and tenth convolutional layer,
while the convolutions are zero-padded in order to preserve size in the frequency dimension.

Along the time dimension the application of pooling is less straightforward.
It was argued in \cite{sainath2013deep} that downsampling in time should be avoided,
rather pooling with stride 1 was used, which does not downsample in time.
However, in \cite{sercu2015very} we did pool with stride 2 in the two top pooling layers.
This fits in the design of our CNNs, where zero-padding is applied
both along the time and frequency directions, so pooling is the only operation
which reduces the context window from its original size (e.g. 16) to the final size (e.g. 4).
This design is directly analogous to VGG networks in computer vision.

\begin{table}[t]
\centering
\begin{tabular}{l | l | l | l | l }
    & \multicolumn{2}{|c|}{SWB-1 (300h)} & \multicolumn{2}{|c}{SWB (2000h)} \\ \hline
                                         & CE   & ST   & CE   & ST   \\ \hline
    Classic 512 CNN \cite{soltau2014joint}   & 13.2 & 11.8 & 12.6 & 10.4 \\  
    Classic+AD+Maxout \cite{saon2015ibm} & 12.6 & 11.2 & 11.7* & 9.9*  \\   
    DNN+RNN+CNN \cite{saon2015ibm}       & -    & -    & 11.1 & \bf{9.4}  \\
    \hline
    (a) Pool                             & 11.8 & \bf{10.5} & 10.2 & \bf{9.4}   \\
    (b) No pool                          & 11.5 & 10.8 & 10.7 & 9.7   \\ 
    (c) No pool, no pad                  & 11.9 & 10.8 & 10.8 & 9.7   \\ 
\end{tabular}
\caption{\label{tab:abc} WER on the SWB part of the Hub5'00 testset,
    for architectures (a), (b) and (c). Column titles show training dataset and method (cross-entropy
    or sequence trained).
    For SWB (2000h) cross-entropy training we initialize with networks that are cross-entropy trained
    on SWB-1 (300h).
    *New results \cite{saon2016ibm}. AD refers to Annealed Dropout.}
    \vspace{-0.3em}
\end{table}

Apart from this practical reason, we did not justify this design choice in \cite{sercu2015very}.
We hypothesize that downsampling in time has both an advantage and a disadvantage.
The advantage is that higher layers in the network are able to access more context and
can learn useful invariants in time.
As argued in \cite{lecun1995convolutional}, once a feature is detected in the lower layers, its exact location
does not matter that much anymore and can be blurred out, as long as its approximate relative position is conserved.
The disadvantage is that the resolution is reduced with which neighboring but different CD states can be distinguished,
which could possibly hurt performance.
In this section we empirically investigate whether pooling in time is justified.

Figure \ref{fig:WDXpWDXnpWDX} summarizes three variations of the 10-layer architecture,
(a) being the original version of \cite{sercu2015very},
Figure \ref{fig:WDXpWDXnpWDX} (b) shows an alternative to pooling in time.
To reduce the context from its original size to the size we want to absorb in the fully connected layer,
we simply omit time-padding on top layers as needed to achieve the desired reduction.

In table \ref{tab:abc} row (a) and (b)
we compare results with and without timepooling.
We see that architecture (a) with time-pooling outperforms architecture (b) after sequence training
and training on 2000 hours.
The result after 2000 hours and sequence training matches the system combination of a
classical CNN, DNN and RNN from \cite{saon2015ibm}, which was the state of the
art system combination result.
Also note that the CE number on SWB (2000h) is far better than the baselines, but the gains from ST are less.
This can be explained by the fact that we do stochastic rather than HF sequence training (see section \ref{sec:trainingdetails}),
which leaves an opportunity for improvement.

From comparing results for model variants (a) with (b),
we conclude that better WER accuracy is achieved by models with
strided pooling along the time dimension.

\section{Efficient convolution over full utterances}
\label{sec:efficient}

\begin{figure}[ht]
    \vspace{-0.3em}
    \centering
    \includegraphics[width=\linewidth]{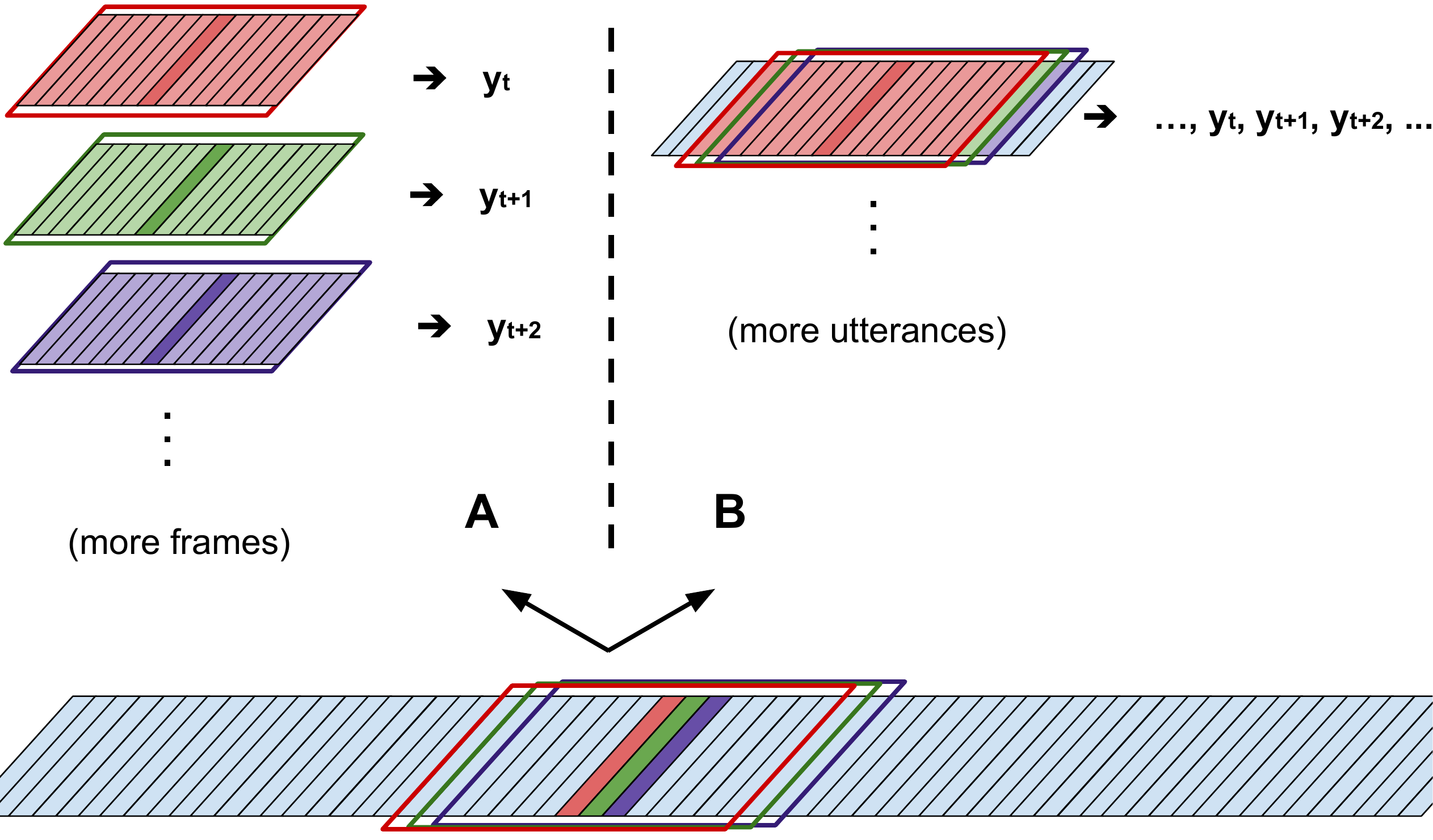}
    \vspace{-0.3em}
    \caption{Two ways of evaluating a full utterance: (A) by splicing as different samples in a minibatch,
        (B) efficient convolutional evaluation, treating the full utterance as a single sample.
        Spliced (A) duplicates the amount of input (and computation) with a factor {\tt context\_size}.
        Efficient convolutional evaluation (B) does not duplicate input and computation,
        but rather takes the full utterance as a single input sample and produces correspondingly large feature maps
        on the intermediate convolutional layers.}
    \label{fig:efficienteval}
    \vspace{-0.3em}
\end{figure}

During sequence training and at test time, it is desirable to process
an utterance at once in a convolutional way as originally described in the foundational CNN papers
\cite{lecun1995convolutional, lecun1998gradient}.
For classical CNNs this has never been a problem.
However, our best performing very deep CNN (figure \ref{fig:WDXpWDXnpWDX} (a)) introduced padding and pooling
along the time dimension, different than prior CNN architectures for LVCSR.
This destroys the desirable property of being able to process a full utterance at once,
outputting a ``dense prediction'', i.e. a CD~state probability for each frame in the utterance:

\bit
\it
When we pool in time with stride of 2, the number of frames for which we have an output is reduced by a factor of 2;
pooling p times (in p layers) results in reduction by a factor $2^p$.
In our first network (a) this means that for an utterance of length \verb$uttlen$ the number
of output frames will be \verb$uttlen / 4$.
This is obviously problematic since the HMM and decoder expect CD~state probabilities for each frame.

\it
The problem with time padding is more subtle.
Consider the first convolutional layer: the zero padding in time makes the edges dependent on the location of the
window. Shifting one frame to predict the next timestep, the values on the edges are 
now truncated by zero padding and have to be recomputed.
On the second convolutional layer, the zero padding changes the outer two edge values,
on the third layer the three outer edge values are modified, etc.

This is illustrated in figure \ref{fig:WDXpWDXnpWDX} (b).
The light red squares at the edges are zero-padding in time. 
The red dashed line indicates which output values in the CNN are modified by the timepadding,
as compared to the same network without timepadding.
The modification travels inwards deeper in the network.
Everything outside the dashed lines is modified by the timepadding 
which is specific to the window location.
\eit

The one obvious way to still obtain an output for each frame for architecture (a)
with pooling and padding in time is what we call ``splicing'' the utterance into \verb$uttlen$ 
different samples in a minibatch, one sample per window for which we want an output.
This is depicted in figure \ref{fig:efficienteval} A,  
with regular full efficient convolution in figure \ref{fig:efficienteval} B.
This way the amount of computation is multiplied by a factor $(1+2\text{ctx})$
when sequence training or testing the network.

\begin{table}[t]
\centering
\begin{tabular}{l | l | l | l}
    Variant             & CE   &  ST  & Full utterance eval?  \\ \hline
    (b) No Pool         & 306  &  207 & Spliced               \\
    (c) No Pool, no pad & 282  &  622 & Efficient convolution \\
\end{tabular}
\caption{\label{tab:fps} Training speed of the very deep CNNs in frames per second (higher is better).
Numbers are average over experiments, for our CuDNNv3-based torch implementation 
executed on a single NVIDIA Tesla K40, including all overhead of data loading and host-device transfer.
Architecture (a) is omitted because the computational cost is very similar to architecture (b).
}
\end{table}

To avoid the need to splice an utterance during sequence training,
we propose the CNN design from figure \ref{fig:WDXpWDXnpWDX} (c),
which does not have any timepadding and pooling, and therefore takes a larger temporal context window.
During sequence training, this CNN model can efficiently convolve over a full utterance.
For (c) the increased context on the lower layers gives a slightly increased
computational cost during CE training (table \ref{tab:fps}, left column).
However architecture (c) is the only architecture that allows for efficient sequence training and deployment
(table \ref{tab:fps}, right column).
The WER results of network (c) is in the bottom row of table \ref{tab:abc}.
The results of (c) are not significantly different from (b) after 300h-ST and 2000h.

As we will discuss in the next section, next to computational efficiency, another advantage to architecture (c) 
is the fact that this architecture allows for a modified version of batch normalization at sequence training time.

\section{Batch Normalization}
\label{sec:bn}
Batch normalization (BN) \cite{ioffe2015batch} is a technique to accelerate training and
improve generalization which gained a lot of traction in the deep learning community.
The idea of BN is to standardize the internal representations inside the network (i.e. the layer outputs), 
which helps the network to converge faster and generalize better, inspired by the way
whitening the network input improves performance.
BN is implemented by standardizing the output of a layer before applying the nonlinearity,
using the local mean and variance computed over the minibatch,
then correcting with a learned variance and bias term ($\gamma$ and $\beta$ resp):

\begin{equation}
    \text{BN}(x) = \gamma \frac{x- E[x]}{(\text{Var}[x]+\epsilon)^{1/2}} + \beta
\end{equation}

The mean and variance (computed over the minibatch) are a cheap and simple-to-implement 
stochastic approximation of the data statistics at this specific layer, for the current
network weights.
Since at test time we want to be able to do inference without presence of a minibatch,
the prescribed method is to accumulate a running average of the mean
and variance during training to be used at test time.
For CNNs, the mean and variance is computed over samples and spatial location.

The standard formulation of BN for CNNs can be readily applied to cross-entropy training during which the
minibatch contains samples from different utterances, different targets and different speakers.
However, during sequence training, spliced evaluation as in figure \ref{fig:efficienteval} A is problematic.
If we construct a minibatch from consecutive windows of the same utterance,
the minibatch mean and variance will be bad for two reasons:
\bit
\it The consecutive samples are identical except the shift and border frame.
    Thus the different samples in the minibatch will be highly correlated.
\it The GPU memory will limit the number of samples in the batch (to around 512 samples on our system).
    Therefore the mean and variance can typically only be computed over few utterances or even just a chunk of an utterance.
\eit
Both these reasons cause the mean and variance estimate to be a poor approximation of the true data statistics,
and to fluctuate strongly between minibatches.

We can drastically improve the mean and variance if we have an architecture that 
allows for efficient convolutional processing of a full utterance like figure \ref{fig:efficienteval} B.
In this case both of the issues are solved: firstly there is no duplication from splicing. 
Secondly, several utterances can be processed in one minibatch, since now they fit in GPU memory.

We aim to maximize the number of frames being processed in a minibatch,
in order for the mean and variance to become a better estimate.
We achieve this by matching the number of utterances in a minibatch with the utterance length, such
that (number~of~utterances) $\times$ (max~utterance~length) $=$ (constant~number~of~frames).
The algorithm for batch assembly can be expressed in pseudo-code as:
\bit
\it choose numFrames to maximize GPU usage
\it while (training):
\bit
   \it targUttLen $\leftarrow$ sample from p(uttLen) $\sim$ f(uttLen) $\times$ uttLen
   \it numUtts    $\leftarrow$ floor(numFrames / uttLen)
   \it minibatch  $\leftarrow$ sample (numUtts) utterances with length close to targUttLen
   \eit
\eit

With our implementation of the 10-layer network of figure \ref{fig:WDXpWDXnpWDX} (c),
and a 12 GB Nvidia Tesla K40 GPU,
we found numFrames = 6000
to be optimal, taking up about 11GB of memory on the device.

\begin{table}[t]
\centering
\begin{tabular}{l | l | l | l | l }
Version (Fig \ref{fig:WDXpWDXnpWDX}) & \multicolumn{2}{|c|}{SWB-1 (300h)} & \multicolumn{2}{|c}{SWB (2000h)} \\ \hline
    & CE          & ST & CE        & ST \\ \hline
(a)      & 11.8  & \bf{10.5} & 10.2 & \bf{9.4}   \\
(b)      & 11.5  & 10.8 & 10.7 & 9.7   \\
(b) + BN & 11.7  & 11.3 & 10.5 & 10.4   \\
(c) & 11.9 & 10.8 & 10.8  & 9.7   \\
(c) + BN & 11.8 & \bf{10.5}  & 10.8 & 9.5 \\
\end{tabular}
\caption{\label{tab:BN} WER on the SWB part of the Hub5'00 testset.
    }
    \vspace{-0.5em}
\end{table}

Table \ref{tab:BN} shows the results of the architectural variants (b) and (c) with and without BN.
As expected, for architecture (b) with batch normalization we do not obtain good performance from sequence 
training, since we have to resort to spliced (inefficient) evaluation.
The performance is worse with than without BN.
In contrast, using the efficient convolutional evaluation with architecture (c), using batch normalization
improves performance from 10.8 to 10.5 on SWB-1 (300h), matching the performance of the superior architecture (a).
On SWB (2000h) adding BN brings the WER down to 9.5, almost matching the result of (a).
We conclude that model (c) with BN recovers the lost performance from model (a).

\section{Training details}
\label{sec:trainingdetails}
We follow the same hybrid setup as \cite{saon2015ibm}, with 32k CD~states (decision tree leaves), 
40D logmel features with window size described in figure \ref{fig:WDXpWDXnpWDX}.
All our work was done using the torch environment~\cite{collobert2011torch7}.
Both CE and sequence training are SGD-based with batch size 128 (except (c) ST as described in section \ref{sec:bn}), 
and $L_2$ weight penalty of $1e^{-6}$.

During CE training, we found SGD (learning rate $0.03$) to work best for networks without BN,
and nesterov accelerated gradient (NAG) with learning rate $0.003$ and momentum $0.99$ for networks with BN.
We use a fixed learning rate decay scheme which divides the learning rate by 3 after 150M, 250M, and 350M frames.
To deal with class imbalance, we adopt the balanced sampling from \cite{sercu2015very}, by sampling
from context dependent state $CD_i$ with probability $p_i = \frac{f_i^\gamma}{\sum_j f_j^\gamma}$.
We keep $\gamma=0.8$ throughout the experiments.

We found two elements essential to make sequence training work well in the stochastic setting:
\bit
\it NAG with momentum 0.99, which we dropped to 0.95 after 100M frames (inspired by~\cite{sutskever2013importance}).
\it Regularization of ST by adding the gradient of cross-entropy loss, as proposed in~\cite{su2013error}.
\eit

\section{Related Work}
CNNs~\cite{lecun1998gradient} have become a dominant approach for solving large-scale machine learning
problems on natural data, for example in computer vision~\cite{krizhevsky2012imagenet,sermanet2013overfeat,farabet2013learning},
speech recognition~\cite{abdel2012applying, sainath2013deep},
and more recently also on character-level text classification~\cite{zhang2015character} 
and language modeling~\cite{kim2015character}.
Very deep nets with small $3\times3$ filters (VGG net) excel not only on ImageNet classification,
but have shown to transfer well to different tasks like neural image captioning~\cite{xu2015show},
object detection~\cite{girshick2015fast},
semantic segmentation~\cite{long2014fully}, etc.

Pooling in time has been applied as early as the work on Time Delay NNs (TDNNs) \cite{waibel1989phoneme}.
Prior work on CNNs for LVCSR has explored non-strided pooling in time~\cite{sainath2014deep},
while in contrast our strided pooling does subsample in time and obtains superior performance.
CNNs with strided pooling in time have been succesfully used for small-footprint keyword spotting~\cite{sainath2015convolutional},
learning filterbanks from raw signal~\cite{palaz2013estimating},
and the end-to-end CTC-based model in~\cite{amodei2015deep}.


Sequence training was introduced to neural network training in \cite{kingsbury2009lattice}.
We performed stochastic sequence training
as opposed to Hessian-Free sequence training \cite{kingsbury2012scalable} 
which was used in our baselines \cite{soltau2014joint, saon2015ibm}.
As mentioned in section \ref{sec:trainingdetails}, we smoothed the ST loss with CE loss as in \cite{su2013error},
and used Nesterov Accelerated Gradient (nag) as optimization method~\cite{sutskever2013importance}.

Batch normalization (BN) was introduced in \cite{ioffe2015batch},
and is closely related to prior work aimed at whitening the activations inside the network \cite{wiesler2014mean}.
BN was shown to improve the Imagenet classification performance in the GoogLeNet 
architecture \cite{szegedy2015rethinking} and residual networks \cite{he2015deep},
which were the top two submissions in 2015 to the ImageNet classification competition.
When applying batch normalization to sequence data,
stacking multiple utterances as one batch for computing
the mean and variance stastistics, is identical to how BN was applied to recurrent neural networks
in \cite{laurent2015batch, amodei2015deep}.

\section{Discussion}
In this paper we demonstrated the strength of very deep convolutional
networks applied to speech recognition in the hybrid NN-HMM framework.
We obtain a WER of 9.4 after sequence training on the 2000 hour switchboard dataset,
which as a single model matches the performance of the state of the art model 
combination DNN+RNN+CNN from \cite{saon2015ibm}.
This model, when combined with a state of the art RNN acoustic model and better language models, 
obtains significantly better performance on Hub5 than any other published model, see \cite{saon2016ibm}.

We compared three model variants, and discussed the importance of time-padding and time-pooling.
\bit
\it Architecture (a) with pooling performs better then (b) and (c) without pooling.
\it Only architecture (c) without padding or pooling allows for batch normalization 
    and efficient convolutional processing of full utterances.
\eit

This naturally raises the question whether we can combine the best of both:
pool in time like architecture (a) and efficient convolutional evaluation like architecture (c).
This is possible with a CNN architecture that does pool but does not pad in time,
and compensates for the lost resolution by either applying $\Delta$ offsets as proposed in \cite{sermanet2013overfeat},
or by using non-strided pooling with sparse kernels along the time dimension, 
an elegant technique independently proposed in both~\cite{li2014highly} and~\cite{ yu2015multi}.
We leave this for future work.

  \newpage
  \eightpt
  \bibliographystyle{IEEEtran}

  \bibliography{refs,morerefs}

\end{document}